\documentclass[letterpaper, 10 pt, journal, twoside]{IEEEtran}
\usepackage{amsmath,amsfonts}
\usepackage{algorithmic}
\usepackage{array}
\usepackage{textcomp}
\usepackage{stfloats}
\usepackage{url}
\usepackage{verbatim}
\usepackage{graphicx}

\usepackage{balance}

\usepackage{tikz}
\usepackage{adjustbox}
\usepackage{subfigure}
\usepackage{chngcntr}
\usepackage{siunitx}
\usepackage{wrapfig}
\usepackage{booktabs}
\usepackage{hyperref}

\begin{document}
\title{A Stack-of-Tasks Approach \\Combined with Behavior Trees:\\ a New Framework for Robot Control}

\author{David C\'{a}ceres Dom\'{i}nguez*, Marco Iannotta*, Johannes A. Stork, Erik Schaffernicht, and Todor Stoyanov%
\thanks{Authors are with the Autonomous Mobile Manipulation Lab at the Center for Applied Autonomous Sensor Systems (AASS), Orebro University, Sweden (e-mail: \href{mailto:david.caceres-dominguez@oru.se}{david.caceres-dominguez@oru.se}; \href{mailto:marco.iannotta@oru.se}{marco.iannotta@oru.se}; \href{mailto:johannesandreas.stork@oru.se}{johannesandreas.stork@oru.se}; \href{mailto:erik.schaffernicht@oru.se}{erik.schaffernicht@oru.se}; \href{mailto:todor.stoyanov@oru.se}{todor.stoyanov@oru.se})}
\thanks{*denotes equal contribution.}
}

\maketitle

\begin{abstract}
Stack-of-Tasks (SoT) control allows a robot to simultaneously fulfill a number of prioritized goals formulated in terms of (in)equality constraints in error space.
Since this approach solves a sequence of Quadratic Programs (QP) at each time-step, without taking into account any temporal state evolution, it is suitable for dealing with local disturbances.
However, its limitation lies in the handling of situations that require non-quadratic objectives to achieve a specific goal, as well as situations where countering the control disturbance would require a locally suboptimal action.
Recent works address this shortcoming by exploiting Finite State Machines (FSMs) to compose the tasks in such a way that the robot does not get stuck in local minima. Nevertheless, the intrinsic trade-off between reactivity and modularity that characterizes FSMs makes them impractical for defining reactive behaviors in dynamic environments.
In this letter, we combine the SoT control strategy with Behavior Trees (BTs), a task switching structure that addresses some of the limitations of the FSMs in terms of reactivity, modularity and re-usability. Experimental results on a Franka Emika Panda 7-DOF manipulator show the robustness of our framework, that allows the robot to benefit from the reactivity of both SoT and BTs.
\end{abstract}

\section{Introduction}
\label{sec:introduction}

\begin{figure}[t!]
\centering
\includegraphics[width=0.97\linewidth]{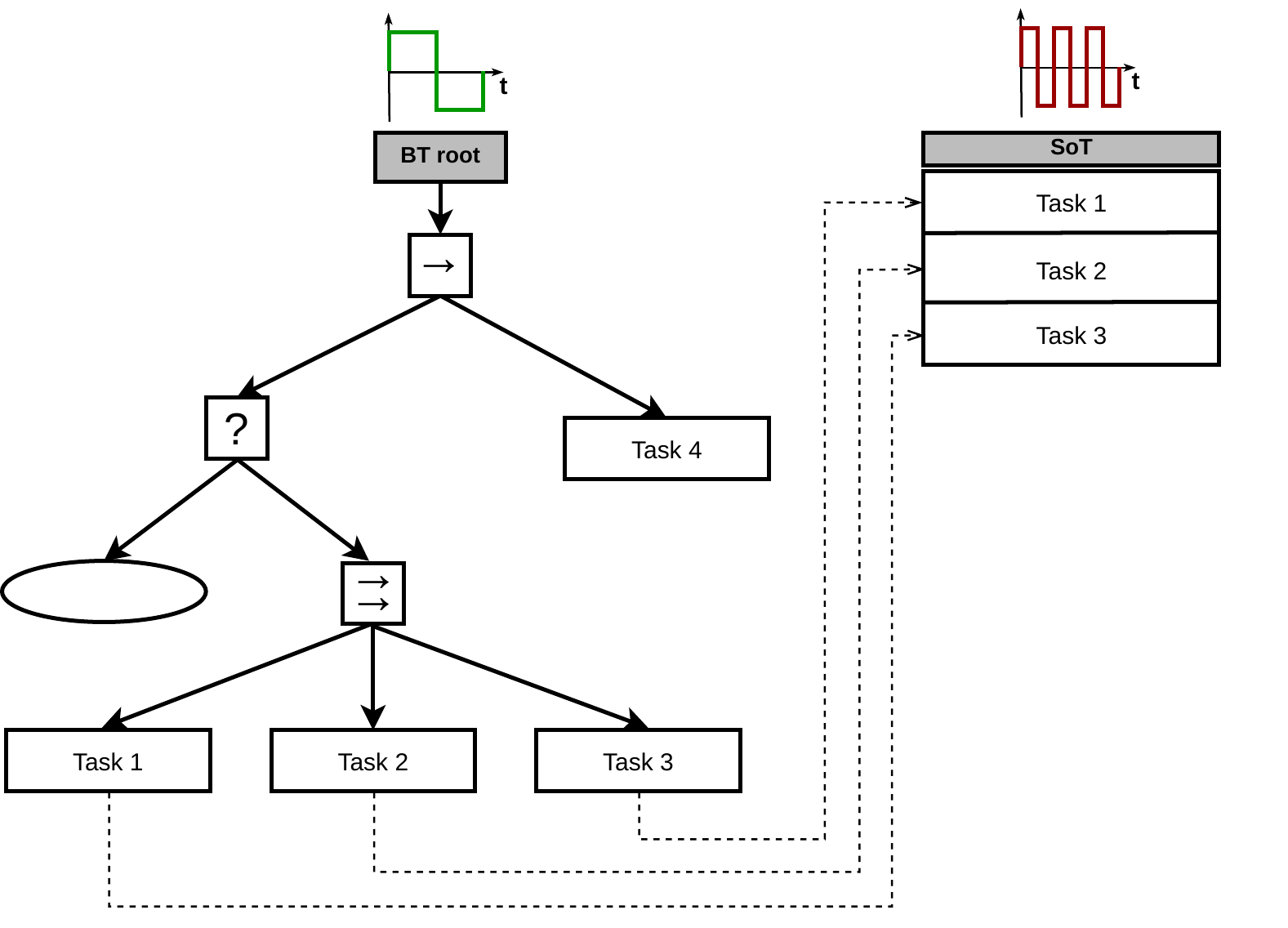}
%
\caption{Integration between the SoT approach and BTs in our framework. The BT and the SoT approach run in parallel at different control frequencies. On the basis of the current robot and environment state, the BT configures a hierarchical control problem, which is then solved by the SoT control strategy at a much higher control frequency. The new tasks are set by the leaf nodes of the BT and the tasks that are no longer needed are removed by the internal nodes. More details are in \ref{subsec:sot-bt}.}
\label{fig:bt-sot} 
\end{figure}

Mechanical systems with a high number of degrees of freedom, such as robot arms or legged robots, can fulfill multiple goals simultaneously.
In the context of robot motion generation, goals can be encoded by means of a \textit{task}, or \textit{error}, function that maps a kinodynamic robot configuration to a scalar value --- e.g., a particular end-effector pose can be formulated as a task via the forward kinematics function of the robot state.
Control of these systems has been addressed by Siciliano et al. \cite{240390} and Kanoun et al. \cite{5766760}, who proposed a local and hierarchical control strategy: \textit{task-priority} or \textit{Stack-of-Tasks} (SoT).
This approach allows to specify a priority for each task, preventing the lower-ranked tasks (e.g., move the end-effector to a position) from interfering with the higher-ranked ones (e.g., avoid collision with a wall).
The SoT approach is local, which allows for fast and efficient implementations that can deal with changes on-line by greedily taking the locally optimal action at every iteration, without having to rely on a typically slower global re-planning.
However, since this control strategy relies on the minimization of quadratic error functions, it can only address problems which can be formulated in terms of quadratic objectives.
This shortcoming can be tackled by using either a motion planning algorithm (e.g., LQR-trees \cite{lqr-trees-tedrake}) or higher-level decision logic, such as Finite State Machines, that can be used to switch between different SoT controllers.
However, both solutions affect the reactivity of the SoT control.
On the one hand, a re-planning of the motion would be needed in case of unexpected changes in the environment.
On the other hand, Finite State Machines (FSMs) allow to compose tasks to avoid local minima, but are characterized by an intrinsic trade-off between reactivity (i.e., ability to quickly and efficiently react to changes) and modularity (i.e., system's  components may be separated into building blocks, and recombined) \cite{DBLP:journals/corr/abs-1709-00084}. This limitation makes them impractical for defining reactive behaviors in dynamic environments.

Recent work has explored Behavior Trees (BTs) as a task switching structure in the context of robotics and artificial intelligence \cite{DBLP:journals/corr/abs-2005-05842}.
Despite BTs being functionally equivalent to FSMs \cite{petter-bts, 6907656}, they promise to address their limitations in reactivity and modularity \cite{DBLP:journals/corr/abs-1709-00084}.
The execution of a BT relies on the generation of a signal, or \textit{tick}, that is propagated from the root to the children. However, the frequency of tick generation is too slow to use BT for direct high-frequency control (i.e., to use them for direct hardware control).
In order to make them usable in this context, a real-time controller in between is typically used, and control functionalities are encapsulated into the leaf nodes of the BT that are responsible for performing commands.
Current works encapsulate entire \textit{skills} (e.g., pushing an object) that typically involve multiple goals in these leaf nodes.
This choice makes the inner working behind each skill not explicitly transparent and can negatively affect the skill re-usability, since some of the specific goals might not be suitable for different contexts.

In this letter, we introduce a new framework for robot control by employing BTs as a task composition structure for an SoT control strategy (Fig. \ref{fig:bt-sot}).
In our approach, a BT and an SoT strategy run in parallel at different control frequencies.
On the basis of the current robot and environment state, the BT periodically configures a hierarchical control problem, which is then solved by the SoT strategy at a much higher control frequency.
The new tasks are set by the leaf nodes of the BT and the tasks that are no longer needed are removed by the internal nodes.

Our main contribution is a novel method to combine a Stack-of-Tasks approach with Behavior Trees in a unified framework, addressing some of the limitations of each model.
On one hand, we are able to extend BTs to high-frequency control tasks, without losing transparency and modularity, since each goal is independently set by a leaf node in the BT.
On the other hand, we endow the SoT approach with higher-level decision logic, without having to compromise on reactivity and modularity.
This makes it easier to manage the situations that have locally optimal solutions, but not globally optimal ones, and facilitates re-usability of sub-behaviors in different scenarios.
Since robotics tasks are, on a high level, typically formulated in terms of sub-tasks that are sequentially or simultaneously executed, our approach provides a transparent and modular tool for the design of robotic behaviors which involve high-frequency control.

\begin{figure}[t]
\vspace{0.3cm}
\centering
\includegraphics[width=0.9\linewidth]{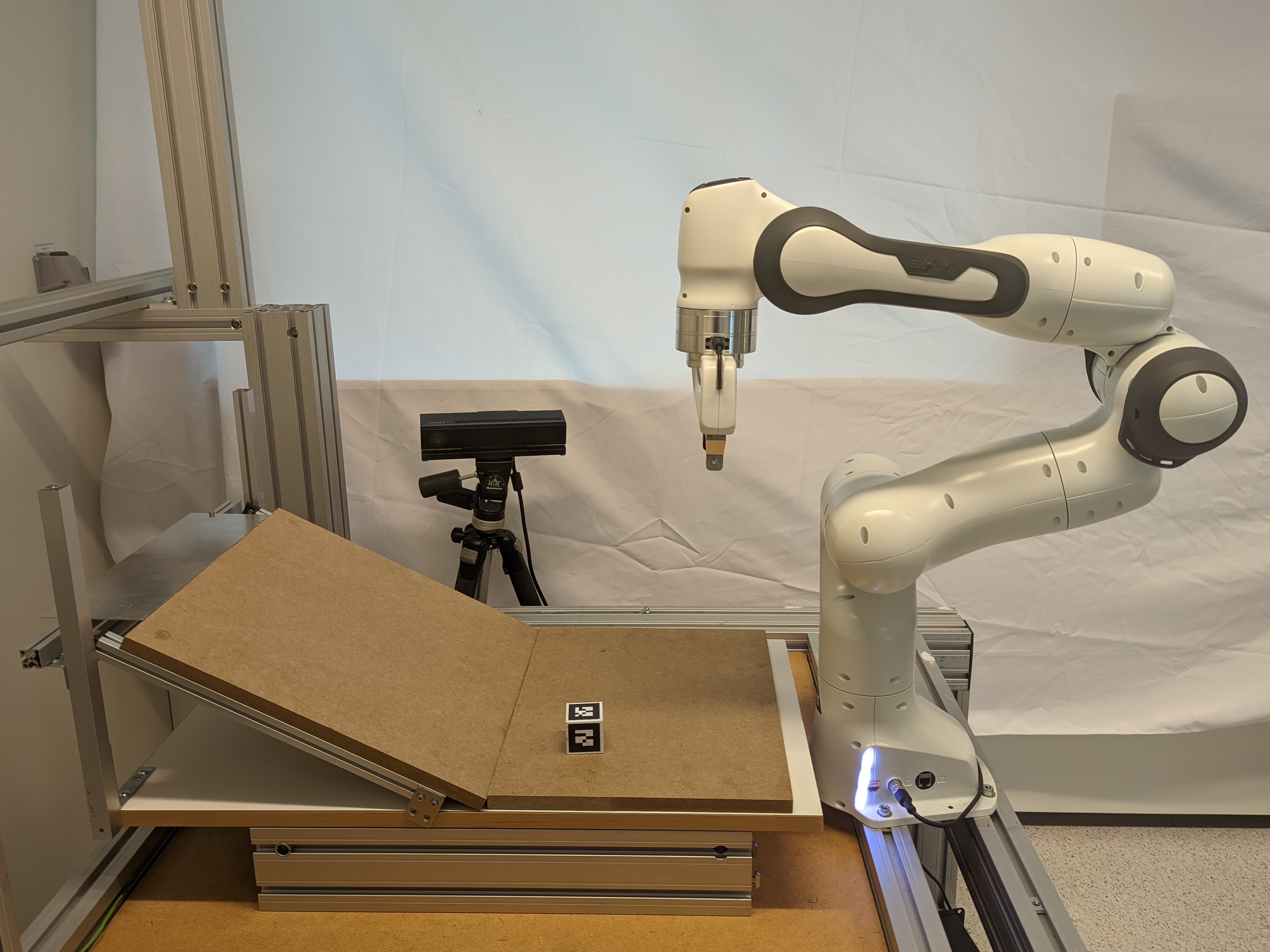}
%
\caption{Experimental set-up. Microsoft Kinect V2, Franka Emika Panda 7-DOF manipulator and 40mm cube. The goal is to pick the cube, place it at the base of the ramp, and then push it to the top.}
\label{fig:exp-setup} 
\end{figure}

We validate our methodology on the set-up shown in Fig. \ref{fig:exp-setup}, where the goal is to pick a small cube, place it at the base of a ramp, and then push it to the top.
The experimental results show that, using our approach, the robot is not only able to robustly achieve the goal at the first attempt, but also to react to unexpected changes, by exploiting the intrinsic reactivity of SoT and BTs.

\section{Related Work}
\label{sec:related_work}

\subsection{Hierarchical Stack-of-Tasks Control Strategy}
Each goal for a redundant mechanical system can be formulated in terms of minimizing a separate \textit{task} function of the robot state $\boldsymbol{q}$, which can be regulated with an ordinary differential equation. In case of multiple tasks, the corresponding equations can be sorted by priority and solved each in the solutions set of higher priority tasks (\textit{task-priority} or \textit{Stack-of-Tasks}).
Kanoun et al. \cite{5766760} proposed a prioritized task-regulation framework based on a sequence of quadratic programs (QP) that generalizes the previous \textit{task-priority} framework \cite{240390} to inequality tasks. In \cite{escande2014hierarchical} a more numerically efficient solution of the same problem was proposed.

Successful applications of this approach have been pointed out in different contexts. \cite{7387707} exploits a hierarchical control strategy for autonomous picking and palletizing, while \cite{stoyanov2018assisted}, \cite{NICOLIS20175672} and \cite{8253809} for single-arm and dual-arm robot teleoperation. Outside the robotic manipulation context, \cite{7803330} presents a control based approach on a whole body control framework combined with hierarchical optimization. 
Moreover, \cite{9568848} and \cite{8968452} show it is also possible to exploit the SoT approach in the context of Reinforcement Learning (RL).

Existing methods that do not involve motion planning typically exploit Finite State Machines (FSMs). In this way tasks can be composed such that the robot does not get stuck in local minima when dealing with multi-step manipulation problems that require non-quadratic objectives.
However, FSMs are characterized by an intrinsic trade-off between reactivity and modularity (more details in Sec. \ref{subsec:sot-fsm-shortcomings}). In this letter we propose to overcome the previous shortcoming by exploiting Behavior Trees.

\subsection{Behavior Trees in Manipulation}

Behavior Trees have been successfully applied in the fields of Robotics and Artificial Intelligence \cite{DBLP:journals/corr/abs-1709-00084, DBLP:journals/corr/abs-2005-05842}.

Thanks to their transparency and modularity, they have been used for robotic arm manipulation tasks. Bagnell et al. \cite{6385888} exploit a BT to model the logical structuring of dexterous manipulation tasks. BTs have also been used within industrial robotics, thanks to their intrinsic reactivity.
In \cite{7140065}, Guerin et al. use BTs to provide non-experts with a set of generalizable skills to visually create pick and place operations. 
Related to their work, \cite{d1fbd409b9ce4ba0b49f3f25509ed2a9} presents an extended system for robot operation (or skill) composition, showing demonstrations in wire-bending and polishing operations. 
\cite{8594319} models skills as reusable motion primitives that are dynamically activated when conditions trigger, enabling the composition of complex behaviours. Extending on their work, \cite{9636292} uses black-box optimizers to learn the parameters of the model skills for a BT policy.
The methods above handle \textit{skills} (i.e., grasping or pushing an object) as integral units, making the inner workings not explicitly transparent and obfuscating the action semantics within each skill.

In the more general context of robot missions that involve a number of desired objectives, some of which need to be taken into account at the same time, Özkahraman et al. \cite{ozkahraman-petter} investigate the combination of BTs with Control Barrier Functions. However, since this approach relies on the definition of a BT where only one leaf node can be run at a time, if multiple tasks have to be performed simultaneously, they have to be encapsulated in a single leaf node.

In contrast, in our framework we exploit an SoT strategy for decomposing a skill into a set of prioritised tasks, whose composition is explicitly and concurrently handled by different leaf nodes of the BT.
To the best of our knowledge, there is no previous work leveraging BTs to dynamically compose prioritised tasks in an SoT control strategy.

\section{Background}
\label{sec:approach}

In this section we briefly describe the used methodologies for the proposed approach. 
We start by describing the SoT control strategy in \ref{subsec:sot}. 
Then we discuss the shortcomings of the SoT framework in combination with Finite State Machines. 
In \ref{subsec:bt} we provide the background of BTs and we explain what makes them suitable for our purpose.

\subsection{Hierarchical Stack-of-Tasks Control Strategy}
\label{subsec:sot}
In this work, we build upon the real-time hierarchical Stack-of-Tasks control strategy proposed in \cite{5766760}. In this section, we briefly revise the concept and refer the reader to the original work for the details.

Let the vectors $\boldsymbol{q}$ and $\dot{\boldsymbol{q}}$ be joint configuration and joint velocity respectively, where $n$ is the number of degrees of freedom. In our work, we are interested in redundant mechanical systems (i.e. $n>6$) that are able to fulfill multiple tasks simultaneously.
According to~\cite{escande2014hierarchical}, we define each task map from the joint space to the operational space via the derivatives of error functions $\boldsymbol{e}(\boldsymbol{q})$.
The task evolution is given by
\begin{align}
\label{evolution}
\dot{\boldsymbol{e}}(\boldsymbol{q})&=\boldsymbol{J}(\boldsymbol{q})\dot{\boldsymbol{q}},
\end{align}
where $\boldsymbol{J}(\boldsymbol{q})=\frac{\partial \boldsymbol{e}(\boldsymbol{q})}{\partial \boldsymbol{q}}$ denotes the task Jacobian.
A desired evolution of (\ref{evolution}) can be imposed via control law $\dot{\boldsymbol{e}}^*(\boldsymbol{q})$, where we use simple P controllers of the form
\begin{equation}
\label{control-law}
\dot{\boldsymbol{e}}^*(\boldsymbol{q}) = -K_p \boldsymbol{e}(\boldsymbol{q}),
\end{equation}
where $K_p$ is a diagonal gain matrix. In case of a single priority level equality task, the following least square Quadratic Program (QP) has to be solved
\begin{equation}
    \arg \min_{\boldsymbol{\dot{q}}} \|\boldsymbol{J} \dot{\boldsymbol{q}} - \dot{\boldsymbol{e}}^*\|^2.
\end{equation}

In order to allow for inequality tasks, we henceforth use a general task formulation with upper bounds
\begin{equation}
\label{inequality}
    \boldsymbol{J} \dot{\boldsymbol{q}} \leq \dot{\boldsymbol{e}}^*(\boldsymbol{q}).
\end{equation}

Escande et al. in \cite{escande2014hierarchical} show how this allows to transcribe lower bounds, double bounds and equalities by reformulating the task.

If the constraint in (\ref{inequality}) is infeasible, the  least squares problem can be formulated introducing a slack variable $\boldsymbol{w}$
\begin{align}
    &\min_{\dot{\boldsymbol{q}}, \boldsymbol{w}} \|\boldsymbol{w}\|^2 \\
    \text{subject to} \quad &\boldsymbol{J}\dot{\boldsymbol{q}} \leq \dot{\boldsymbol{e}}^*+\boldsymbol{w}. \nonumber 
\end{align}

To form a hierarchical Stack-of-Tasks with $p=1,\dots,P$ priority levels, we stack all task Jacobians of equal priority $p$ in a matrix $\boldsymbol{A}_p$ and the corresponding upper bounds (i.e., reference velocities) in a vector $\boldsymbol{b}_p$ to form one constraint of the form $\boldsymbol{A}_p\dot{\boldsymbol{q}} \leq \boldsymbol{b}_p$ for each hierarchy level.
The main idea of a hierarchical Stack-of-Tasks control strategy is to solve the tasks of lower priority in the null space of ones of higher priority. Therefore, the following QP needs to be solved for $p=1,\dots,P$
\begin{align}
\label{objective}
    &\min_{\dot{\boldsymbol{q}}, \boldsymbol{w}_p} \|\boldsymbol{w}_p\|\\
    \text{subject to} \quad &\boldsymbol{A}_i \dot{\boldsymbol{q}} \leq \boldsymbol{b}_i+\boldsymbol{w}^*_i, \quad i=1,\dots,p-1 \nonumber \\
    &\boldsymbol{A}_p \dot{\boldsymbol{q}} \leq \boldsymbol{b}_p+\boldsymbol{w}_p, \nonumber
\end{align}
where the previous slack variable solutions $\boldsymbol{w}^*_i$ are frozen between iterations.

This method essentially provides a global control policy by solving a sequence of instantaneous optimal control problems at each time step.

\subsection{Limitations of SoT with Finite State Machines}
\label{subsec:sot-fsm-shortcomings}
The hierarchical SoT control strategy described in \ref{subsec:sot} relies on the assumption that it is always possible to reach a globally optimal solution by minimizing a quadratic error function.
In reality, this is not always the case and it often happens to deal with problems that require non-quadratic objectives.
In these situations, a single SoT might not be enough, because the control strategy might bring the robot to a disadvantageous configuration with respect to the global goal.
The only way to address these scenarios, without considering a more complex error function or a motion planner, is to compose a sequence of local approximations (i.e., Stack-of-Tasks) that prevent the control strategy from falling into a local minimum.

Finite State Machines (FSMs) have been typically used for this purpose, since they are a standard choice in the context of task switching structures. However, as pointed out in \cite{DBLP:journals/corr/abs-1709-00084}, FSMs are characterized by an intrinsic trade-off between reactivity (i.e., ability to quickly and efficiently react to changes) and modularity (i.e., system's  components may be separated into building blocks, and recombined). The state transitions in FSMs are \textit{one-way control transfers}, in the sense that, making an analogy with programming languages, they behave like \textit{Goto} statements, which have well-known shortcomings \cite{10.1145/362929.362947}. Making a system reactive by using FSMs requires defining many transitions (i.e., \textit{one-way control transfers}) between components. As a consequence, this negatively affects the modularity, because if, for example, one component is removed, every transition to that component needs to be revised.

\subsection{Behaviour Trees}
\label{subsec:bt}

\begin{figure}[t!]
\centering

\subfigure[]{\includegraphics[height=0.15\linewidth]{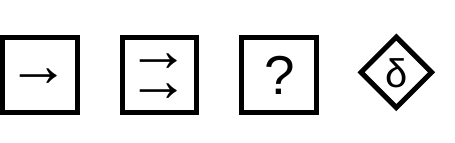}
\label{fig:control-nodes}
}

\subfigure[]{\includegraphics[width=0.5\linewidth]{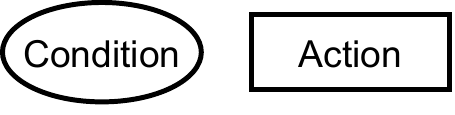}
\label{fig:execution-nodes}
}
\caption{(a) Control flow nodes, from left to right: Sequence, Parallel, Fallback and Decorator. (b) Execution nodes, from left to right: Condition and Action.}
\end{figure}

Behaviour Trees are an alternative to FSMs that promise to address some of the limitations in modularity, re-usability and reactivity.  Following the terminology from \cite{DBLP:journals/corr/abs-1709-00084}, a BT is composed of a root node, and at least one internal node and one leaf node. The execution of a BT starts at the root node, which generates signals, or \textit{ticks}, with a given frequency, that are propagated to the children, down to the leaf nodes. A node is executed if and only if it receives a tick, and returns a status to the parent. The possible statuses are \textit{Running}, if the execution is under way, \textit{Success} if the node has achieved its goal, or \textit{Failure} otherwise.

Internal Nodes, or Control Nodes, can be of type \textit{Sequence}, \textit{Fallback}, \textit{Parallel},  or \textit{Decorator} (Fig. \ref{fig:control-nodes}). A Sequence Node propagates the tick to its children nodes from left to right and it returns \textit{Success} if and only if all its children return \text{Success}. A Fallback Node propagates the tick to its children nodes from left to right until one of them returns \textit{Success}. Note that, when a child returns \textit{Running}, the Sequence and the Fallback Nodes do not tick the next child if any. A Parallel Node propagates the tick to all its children simultaneously and it returns \textit{Success} if a set amount of them return \textit{Success}. Lastly, a Decorator node modifies the return status of a single child node with any custom policy.

Leaf Nodes, or Execution Nodes, include two categories (Fig. \ref{fig:execution-nodes}). When ticked, an Action Node performs a command. It returns \textit{Success} if the action defined by the command is correctly completed, \textit{Failure} if it has failed and \textit{Running} if it is ongoing. A Condition Node checks a proposition, returning \textit{Success} if it holds, \textit{Failure} otherwise.

In the previous section we pointed out that the state transitions in FSMs are \textit{one-way control transfer}. Although BTs are equivalent to FSMs (\cite{petter-bts, 6907656}), they exploit \textit{two-way control transfers}. 
When getting the tick from the parent, a node is executed and returns a status.
As a consequence, the flow of a BT is governed by its internal nodes, on the basis of the status returned by the children.
This substantial difference makes BTs more reactive and modular at the same time when compared to FSMs.
On the one hand, the reactivity is given both by the continual generation of ticks and by the their tree traversal in a closed loop execution. On the other hand, the modularity derives from the possibility of having each sub-tree of a BT as an independent module, with a standard interface given by the return statuses.

The combination of these two aspects makes BTs more suitable than FSMs as a task switching structure for the SoT control strategy.

\section{Proposed Approach}

The main contribution of our letter is the integration of the SoT control strategy with BTs, such that a BT is used to configure the hierarchical control problem to be solved by the SoT approach. 
In \ref{subsec:sot-bt} we explain our framework and in \ref{subsec:example} we provide an example to clarify its functioning.

\subsection{Combining SoT with BTs}
\label{subsec:sot-bt}

\begin{figure}[t!]
\centering

\subfigure[]{\includegraphics[height = 0.15\linewidth]{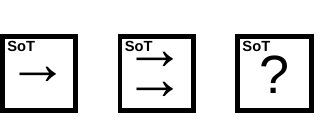}
\label{fig:new-control-nodes}
}

\subfigure[]{\includegraphics[height = 0.11\linewidth]{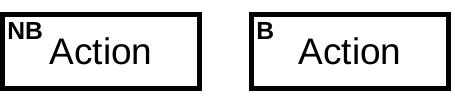}
\label{fig:new-execution-nodes}
}
\caption{New BT nodes in our framework. Nodes in Fig. \ref{fig:new-control-nodes} are \textit{SoT-Control Nodes} which upon completion remove the tasks posed by their children from the SoT tasks set by the children.
Fig. \ref{fig:new-execution-nodes} shows the two new leaf nodes we propose, namely the \textit{Non-Blocking} and \textit{Blocking Action Nodes} respectively. The \textit{Non-Blocking Action Node} sets one task for the SoT control strategy and then returns \textit{Success}. The \textit{Blocking Action Node} sets one task and then returns \textit{Running} until the task error $\boldsymbol{e}(\boldsymbol{q})$ either satisfies the imposed constraint (returns Success) or a timeout occurs (returns Failure).}
\label{fig:new-nodes}
\end{figure}

As mentioned in \ref{subsec:sot-fsm-shortcomings}, a single SoT control instance might not be sufficient in all situations when trying to fulfill a goal.
The aim is to exploit the flow of a BT to dynamically update the task functions in the SoT, while preserving the framework's transparency and modularity.

The basic idea is to have a BT and an SoT approach running in parallel at different control frequencies, and to use the BT for mapping a robot and environment state to a hierarchical control problem. Once configured by the BT, this control problem is then solved by the SoT strategy at a much higher control frequency. In this way, we can constantly update the task functions in the SoT by either setting new tasks, or removing the ones which are no longer necessary.

We exploit Action and Control Nodes in the BT to configure the hierarchical control problem.
Regarding the Action Nodes, we distinguish between \textit{Non-Blocking} and \textit{Blocking} Action Nodes (Fig. \ref{fig:new-execution-nodes}).

A \textit{Non-Blocking} Action Node sets a new task $x = (\boldsymbol{e}_x(\boldsymbol{q}), p_x, {K_p}_x)$, where $\boldsymbol{e}_x(\boldsymbol{q})$ is the \textit{task error function}, $p_x \in \mathbb{N}_+$ is the \textit{task priority}, and ${K_p}_x \in \mathbb{R}_+$ is the \textit{task gain}. As a consequence, a new \textit{task constraint} in the general form of $\boldsymbol{J}_x \dot{\boldsymbol{q}} \leq \dot{\boldsymbol{e}}_x^*(\boldsymbol{q}) + \boldsymbol{w}$ is added to the hierarchical control problem (i.e., the task $x$ has been \textit{set}), where $\dot{\boldsymbol{e}}_x^*(\boldsymbol{q}) = -{K_p}_x \boldsymbol{e}_x(\boldsymbol{q})$ according to Eq. (\ref{control-law}). This Action Node returns \textit{Success} immediately after the task $x$ is set.

A \textit{Blocking} Action Node sets a new task $x = (\boldsymbol{e}_x(\boldsymbol{q}), p_x, {K_p}_x, s_x, f_x)$, where $\boldsymbol{e}_x(\boldsymbol{q}), p_x, {K_p}_x$ are defined as before, while $s_x, f_x, t_x \in \mathbb{R}_+$ are an \textit{error threshold}, a \textit{time threshold} and the \textit{task execution time} respectively. This Action Node returns: 

\begin{itemize}
\item \textit{Success}, if $x$ is set, $\boldsymbol{e}_x(\boldsymbol{q}) \leq s_x$ and $t_x \leq f_x$;
\item \textit{Running}, if $x$ is set, $\boldsymbol{e}_x(\boldsymbol{q}) > s_x$ and $t_x \leq f_x$;
\item \textit{Failure}, if $x$ is not set or $t_x > f_x$.
\end{itemize}

\begin{figure*}[t!hb]
\centering

\subfigure[]{\includegraphics[height = 0.27\linewidth]{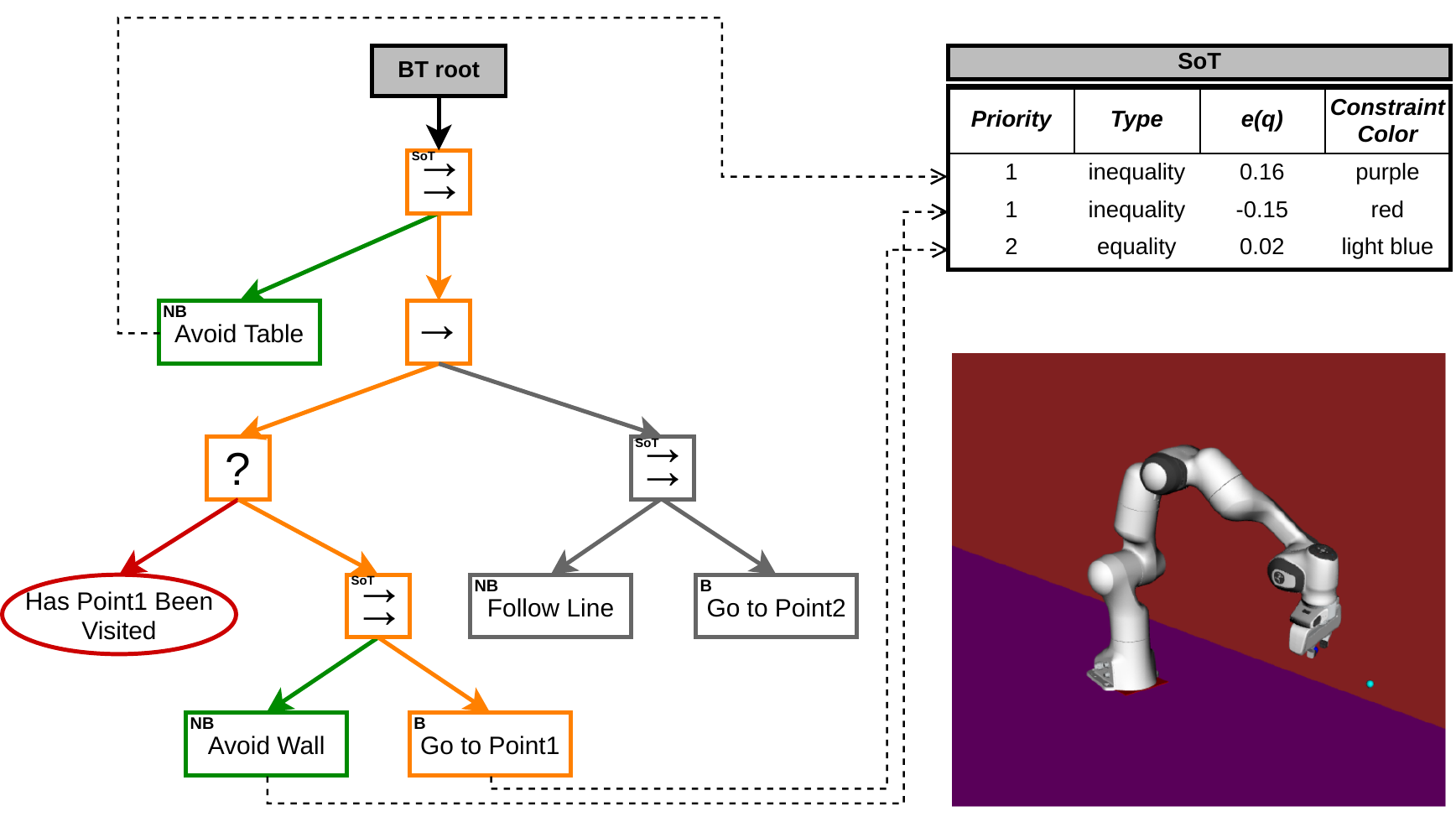}
\label{fig:example-a}
}
\subfigure[]{\includegraphics[height = 0.27\linewidth]{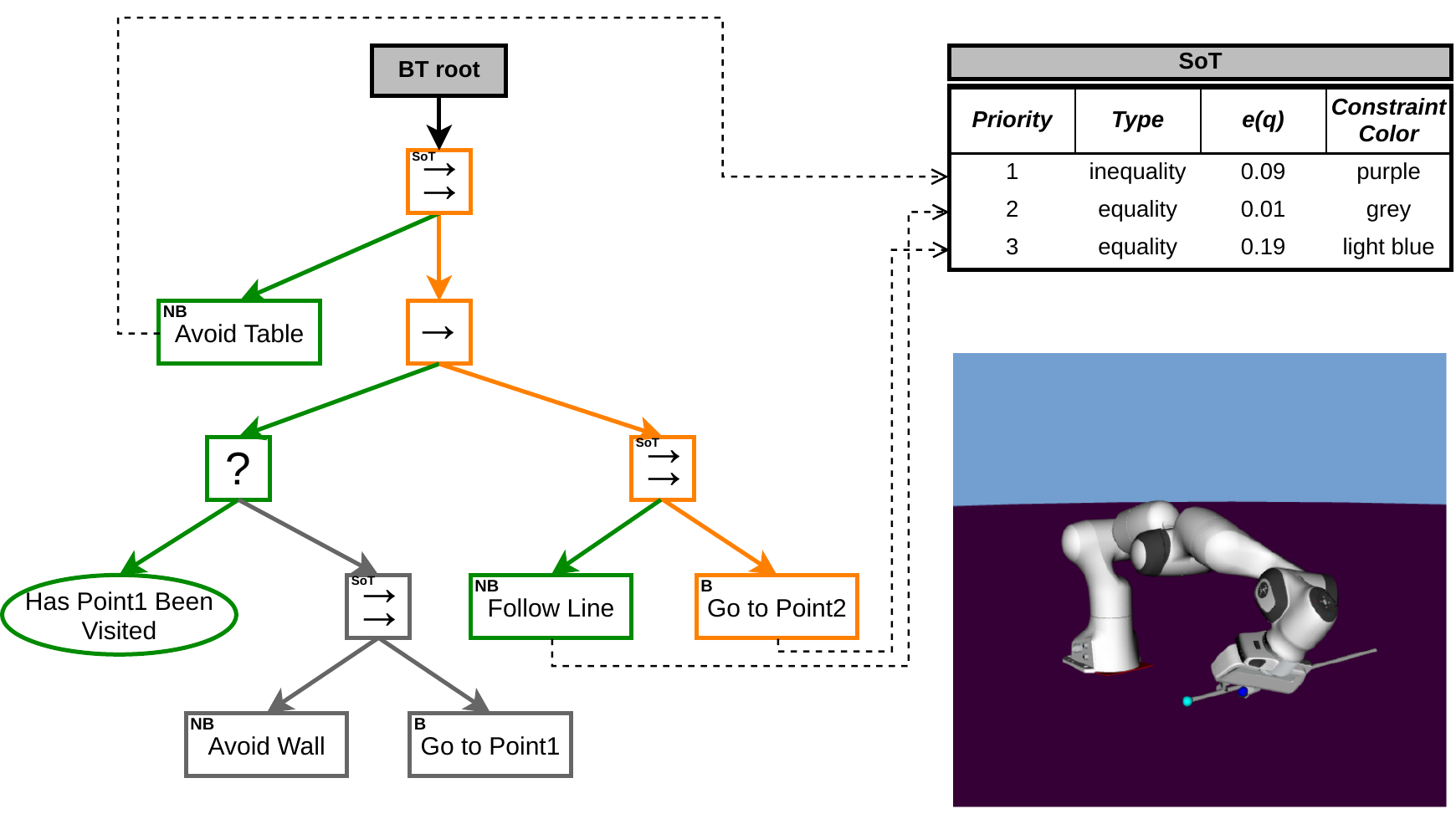}
\label{fig:example-b}
}
%
%
\caption{Functioning example of our framework. The goal for the robot manipulator is to reach two different points, while not violating some constraints. Fig. \ref{fig:example-a} and \ref{fig:example-b} illustrate the robot performing the motions respectively to the first and the second point. For each of them, we show a frame of the running BT (green denotes \textit{Success}, orange denotes \textit{Running}, red denotes \textit{Failure}, and grey denotes \textit{not ticked}), a table with information about the active tasks in the SoT, and a visualization of the obstacle planes, the goal points, and the constraint line associated to the active tasks.}
\label{fig:example}
\end{figure*}

The main difference between the two Action Nodes lies in the \textit{blocking} of the tick, which takes place only in the \textit{Blocking} Action Node.
We need to block the tick of the BT (i.e., return \textit{Running} and avoid to tick the rest of the BT) after a certain control problem is configured by the Action Nodes, in order to wait for reaching a specific robot and environment state before updating the SoT.
Since a \textit{Non-Blocking} Action Node returns \textit{Success} immediately after having set a task, it never returns \textit{Running} and, consequently, never prevents the tick from being propagated to the rest of the tree.
Although it is possible to exploit Condition Nodes for this purpose, their use would imply that no Action Node has an internal mechanism to determine its progress, that the \textit{Running} status is never returned and that the condition is checked at the low frequency of the BT. For these reasons, we introduce the \textit{Blocking} Action Node and exploit two threshold values in it.
On the one hand, we monitor the error function $\boldsymbol{e}(\boldsymbol{q})$ of a task and compare it with an \textit{error threshold} $s_x$, to determine if the task is \textit{completed} (i.e., $\boldsymbol{e}_x(\boldsymbol{q}) \leq s_x$) and, consequently, if the Action Node that sets the task is still \textit{Running} or not.
On the other hand, we monitor the \textit{execution time} $t_x$ of the task and compare it with a \textit{time threshold} $f_x$, to identify possible situations in which the task can not be achieved (i.e., $t_x > f_x$).

The idea of task monitoring is, however, not suitable for all the types of tasks that can be set.
For tasks that define desired motions (e.g., reach a point), a lower value of $\boldsymbol{e}(\boldsymbol{q})$ certainly states that the motion is closer to \textit{completion} (e.g., the robot has almost reached the point).
On the contrary, when taking into account tasks that define constraints (e.g., avoid collision with a wall), the value of $\boldsymbol{e}(\boldsymbol{q})$ can no longer be interpreted in the same way.
For this reason, we use both \textit{Non-Blocking} and \textit{Blocking} Action Nodes in our framework, in order to exploit the first ones for constraint tasks and the second ones for motion tasks.

Setting one task in each Action Node, instead of an entire Stack-of-Tasks, brings two main advantages.
On the one hand, it improves understandability, since the BT provides a clear way both to visualize all the set tasks and to identify the \textit{blocking} ones.
On the other hand, tasks that are in common for an entire subtree and not strictly related to the behavior defined by it (typically constraints) can be defined only once, outside the scope of the subtree.
This makes each subtree more independent and easily re-usable in different contexts.

Regarding the Control Nodes, we extend them to remove the tasks that are no longer needed by creating \textit{SoT}-Control Nodes (Fig. \ref{fig:new-control-nodes}).
Let an \textit{SoT}-Control Node $C$ be a Control Node with $m$ children, of which $n \leq m$ are Action Nodes.
$C$ removes a set of tasks $X_C$, where $x \in X_C$ has been set by an Action Node that is a child of $C$ ($|X_C| = n$).
As a consequence, all the associated \textit{task constraints} previously added by the children of $C$ are removed from the hierarchical control problem.
Task removal takes place immediately before the Control Node returns its final outcome (i.e., \textit{Success} or \textit{Failure}), without affecting the node behavior described in Section \ref{subsec:bt}.
In case of the \textit{Running} status, the task removal takes place if and only if the \textit{SoT}-Control Node is \textit{halted} (i.e., the execution of its children is interrupted), because a previous condition in the BT is no longer met.
The creation of new Control Nodes allows us to exploit these nodes for task removal when needed, without removing the possibility to use the standard ones just for controlling the flow of the BT.

Control Nodes are more suitable than Action and Condition Nodes for removing tasks. On the one hand, task removal in the Action Nodes would make it problematic to remove the tasks set by \textit{Non-Blocking} Action Nodes, since they have to be removed after the tasks set by \textit{Blocking} Action Nodes.
On the other hand, since Condition Nodes have no information about the other nodes in the tree, it would be complex to deduce which tasks have to be removed.
On the contrary, Control Nodes have direct access to their children and provide us with a straight way to specify the tasks to remove by exploiting the BT design.

As a result of the above definitions of the new nodes, the hierarchical control problem is dynamically updated at each tick according to the logic modelled by the BT, and solved in parallel by the SoT strategy according to Eq. (\ref{objective}).

\subsection{Framework Example}
\label{subsec:example}

To clarify the functioning of our framework, we provide a simple illustrative example in Fig. \ref{fig:example}. Consider a robot manipulator on a flat surface, such as a table, near an obstruction, such as a wall. The manipulator must move its end-effector to two successive points. The motion from the start position to the first point requires operating near the obstruction, so collision with it has to be avoided. On the contrary the motion from the first point to the second must be accomplished by keeping the end-effector on a line and it is supposed to occur further away from the obstruction. While performing both motions, the collision avoidance with the table has to be also taken into account.

Fig. \ref{fig:example-a} and \ref{fig:example-b} illustrate the robot performing the motions respectively to the first and the second point. For each of them, we show a frame of the running BT (green denotes \textit{Success}, orange denotes \textit{Running}, red denotes \textit{Failure}, and grey denotes \textit{not ticked}), a table with information about the active tasks in the SoT, and a visualization of the the obstacle planes, the goal points, and the constraint line associated with the active tasks. 

Since the first Control Node under the root is an SoT-Parallel Node, the Action Node \textit{Avoid Table} and the Condition Node \textit{Has Point1 Been Visited} are ticked at the same time.
The Action Node sets an inequality task and returns \textit{Success}\footnote{Note that the task priority is internally defined by the Action Node and not explicitly visible in the BT.}, while the Condition Node returns \textit{Failure}, since the condition is not met.
Then, the SoT-Parallel Node under the Fallback Node ticks \textit{Avoid Wall} and \textit{Go to Point1} simultaneously (Fig. \ref{fig:example-a}).
Similarly to \textit{Avoid Table}, \textit{Avoid Wall} just sets an inequality task and returns \textit{Success}.
The obstacle planes associated to the inequality tasks are represented respectively in purple and red. On the contrary, since \textit{Go to Point1} is a Blocking Action Node, after it sets the equality task, it returns \textit{Running} until the error function of its task ($\boldsymbol{e}(\boldsymbol{q})$ in the table) is lower than the specified threshold.
The goal point is shown in light blue.
Once \textit{Go to Point1} returns \textit{Success}, the SoT-Parallel Node above it removes the tasks previously set by the children and returns \textit{Success}.
As a consequence, the Fallback Node returns \textit{Success} and the Sequence Node above propagates the tick to the next SoT-Parallel Node, which in turn ticks the Action Nodes \textit{Follow Line} and \textit{Go to Point2}.
The former sets an equality task whose constraint line is shown in grey and then returns \textit{Success}, while the latter behaves analogously to \textit{Go to Point1}.
At the next tick, the Condition Node \textit{Has Point1 Been Visited} is met and the Action Nodes \textit{Avoid Wall} and \textit{Go to Point1} are not executed again (Fig. \ref{fig:example-b}).
Once \textit{Go to Point2} returns \textit{Success}, the SoT-Parallel Node above it removes the two tasks set by the children.
Then, the Sequence Node returns \textit{Success} and the tick is back-propagated to the above SoT-Parallel Node, which removes the remaining active tasks (only the one set by \textit{Avoid Table}) and lastly returns \textit{Success}.

\section{Evaluation}
\label{sec:evaluation}

We evaluate our approach with a Franka Emika Panda 7-DOF manipulator.
We build upon the implementation of SoT exploited in \cite{7387707} and the \textit{BehaviorTree.CPP\footnote{https://www.behaviortree.dev/}} library for BTs. 

To test the validity of our framework, we utilize the set-up in Fig. \ref{fig:exp-setup} for real-world experiments and design a task that consists of picking up a 40 mm cube, placing it at the base of the ramp and pushing it up until the cube safely reaches the elevated platform.
The use of a standard two-finger gripper makes the goal challenging to achieve, as it is prone to failure during the pushing operation.
To track the position of the cube, we use a Microsoft Kinect V2 sensor and fiducial markers attached to the surface of the cube. 

The modular nature of BTs allows us to design the BT and tune the controller parameters in isolation for the picking, placing and pushing operations.
Then we combine the sub-trees into a unified BT and evaluate the full system during the experiments.
We tick the BT continuously until either the end goal is achieved (i.e., the BT root node returns \textit{Success} and the cube is at the top of the ramp) or the robot encounters failure scenarios from which it cannot recover (i.e., collisions with the environment).
Regarding BT frequency, the tree is ticked again as soon as the root returns the status of the previous tick. 
This resulted in a variable frequency, as it depends on the number of nodes that need to be ticked before returning a status. 
Moreover, the SoT frequency is also variable, as it depends on the number of tasks that need to be solved for the current stack.
During the experiments, the BT frequency ranged between 2-110 Hz and the SoT frequency ranged between 220-1.500 Hz.

\subsection{Robustness}
\label{sec:robustness}

To show the robustness of our framework, we evaluate the success rate of the previously described operation over 50 trials, considering 5 different fixed starting positions of the cube.

The results are shown in Table \ref{tab:tab1}.
\textit{Attempt 1} refers to the success rate without the BT ever having returned \textit{Failure} (i.e., first attempt).
\textit{Attempt 2} refers to the success rate when the robot has returned one \textit{Failure} and has recovered from it (i.e., second attempt).
\textit{Overall} refers to the success rate at any attempt (i.e., in our case with at most one \textit{Failure}).

In 90\% of the trials, the task was completed at first attempt.
In two instances, the cube fell down the ramp during the pushing. In both cases the system automatically recovered from the failure and succeeded at the second attempt (100\%).

The overall success rate is 94\%. 
The remaining 6\% fail rate is associated to errors during the pose estimation of the fiducial marker on the cube.
This caused a displacement of the end-effector during the picking and resulted in a collision of the cube with the table during the placing, from which the system could not recover.

\begin{table}[h!]
\centering
\caption{Robustness experiments}
\label{tab:tab1}
\begin{tabular}{rrr|rr}
\toprule
Position       & \# Trials   & Overall & Attempt 1  & Attempt 2  \\
\midrule
1              & 10          & 90\%            & 80\%            & 100\%           \\
2              & 10          & 100\%           & 90\%            & 100\%           \\
3              & 10          & 80\%            & 80\%            & -               \\
4              & 10          & 100\%           & 100\%           & -               \\
5              & 10          & 100\%           & 100\%           & -               \\
\midrule
\textbf{Total} & \textbf{50} & \textbf{94\%}   & \textbf{90\%}   & \textbf{100\%} \\
\bottomrule
\end{tabular}
\end{table}

\subsection{Reactivity}

To test the reactivity benefit from combining the SoT approach with BTs, we devise two experiments that simulate different types of disturbance during execution.

\begin{figure}[h!]
\vspace{0.3cm}
\centering

\subfigure[]{
\includegraphics[height = 0.23\linewidth]{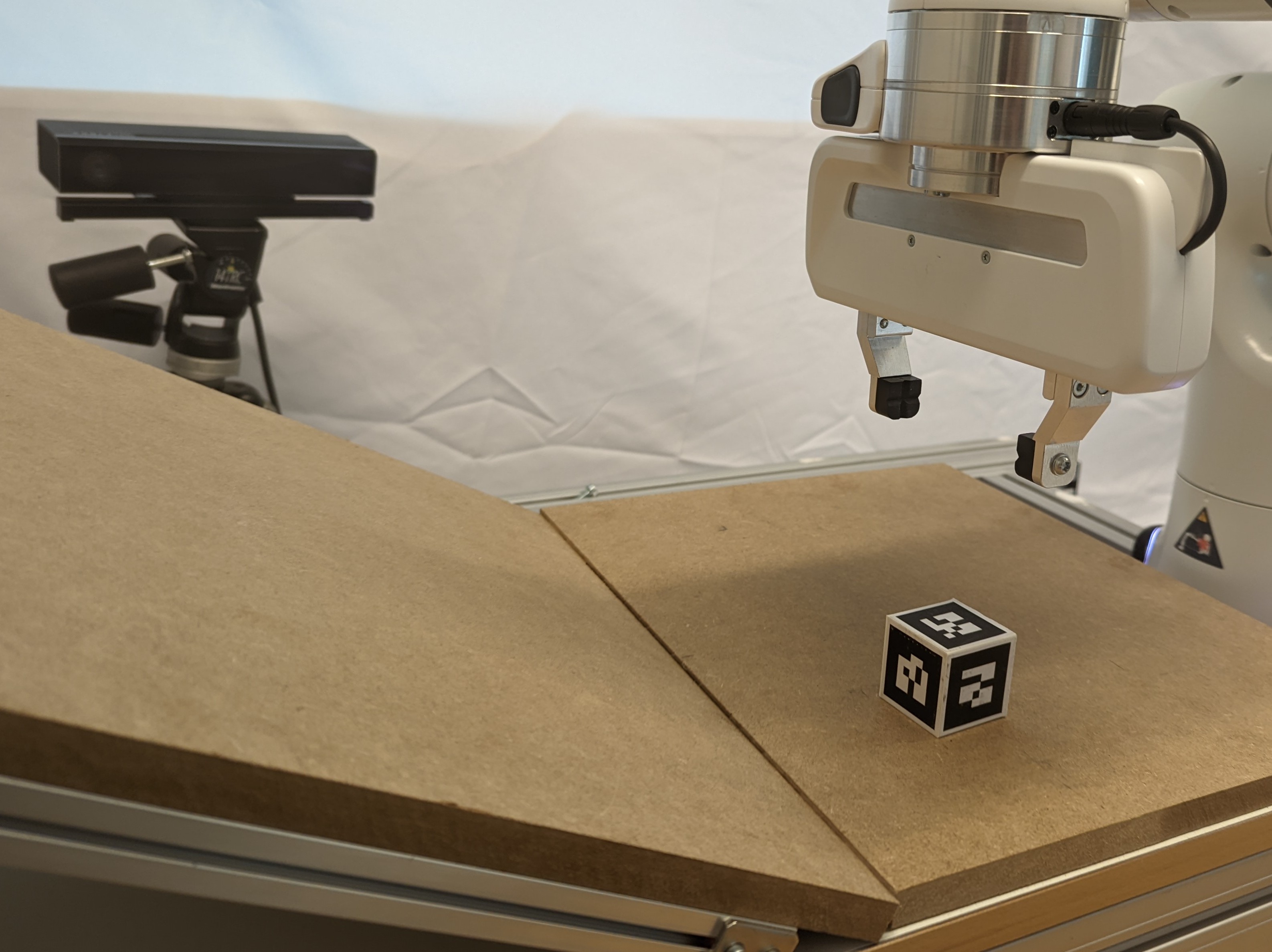} \hspace{0.025cm}
\includegraphics[height = 0.23\linewidth]{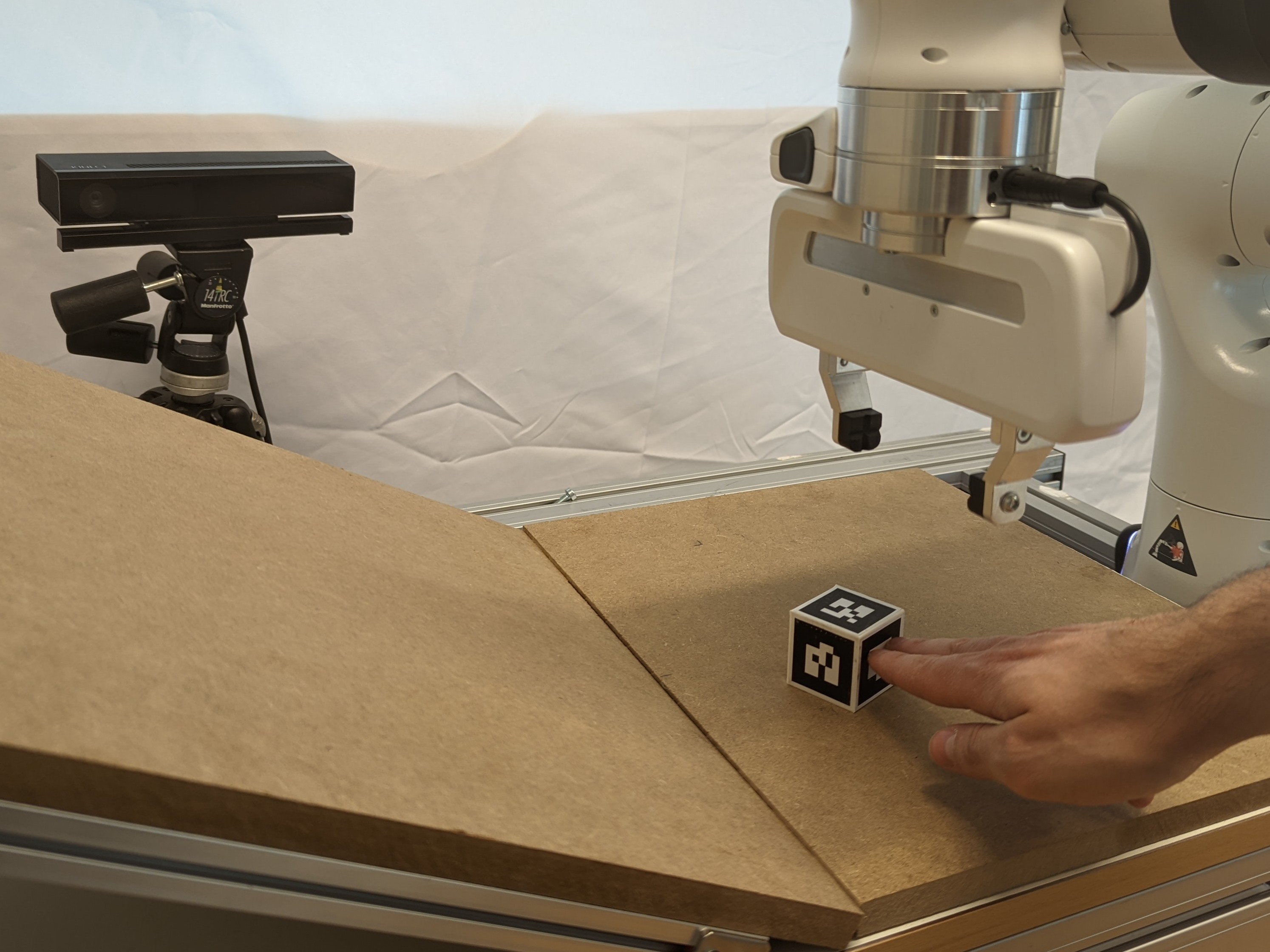} \hspace{0.025cm}
\includegraphics[height = 0.23\linewidth]{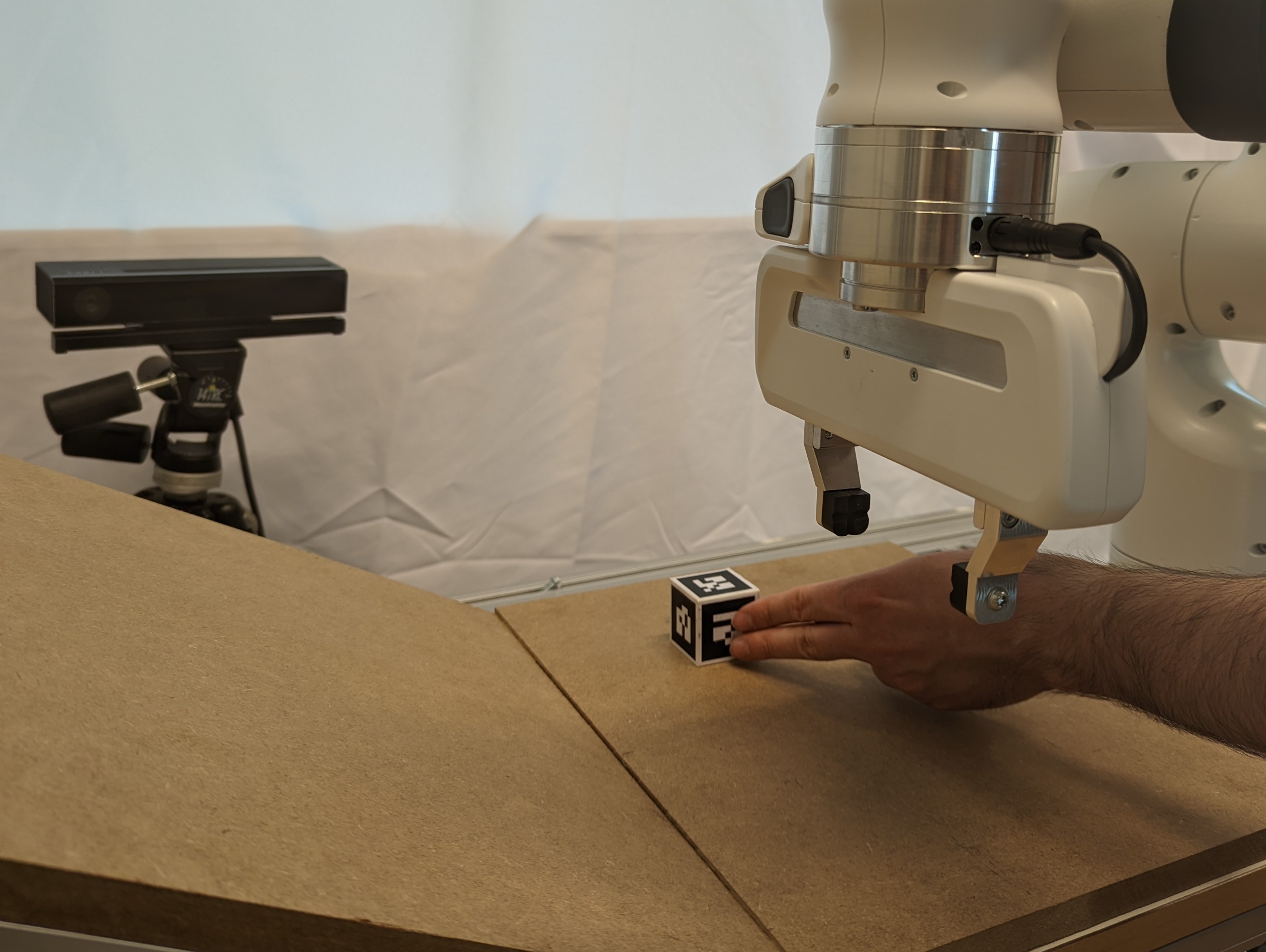} \hspace{0.025cm}
\label{fig:exp-react-local} 
	}

\subfigure[]{
\includegraphics[height = 0.23\linewidth]{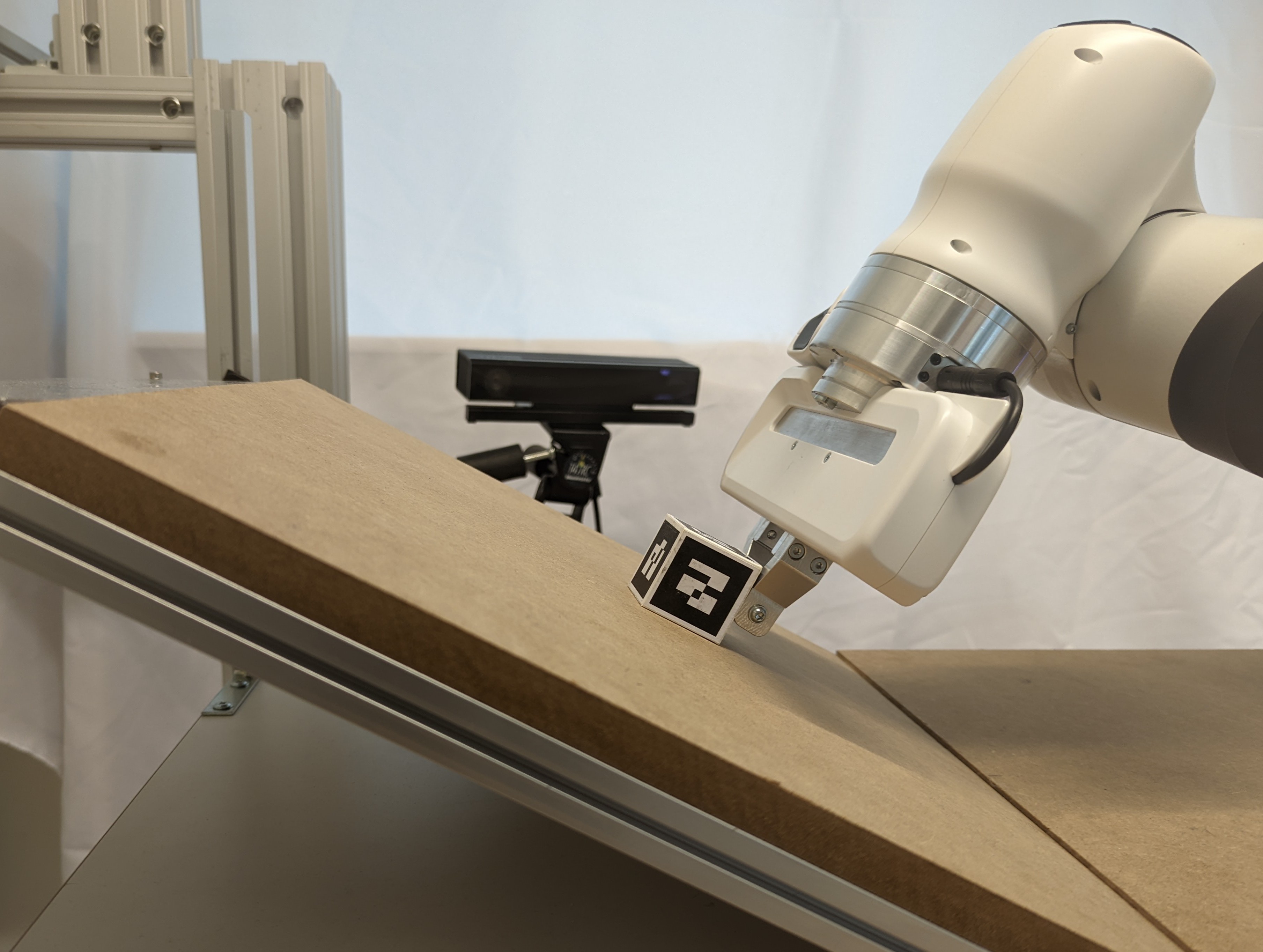} \hspace{0.025cm}
\includegraphics[height = 0.23\linewidth]{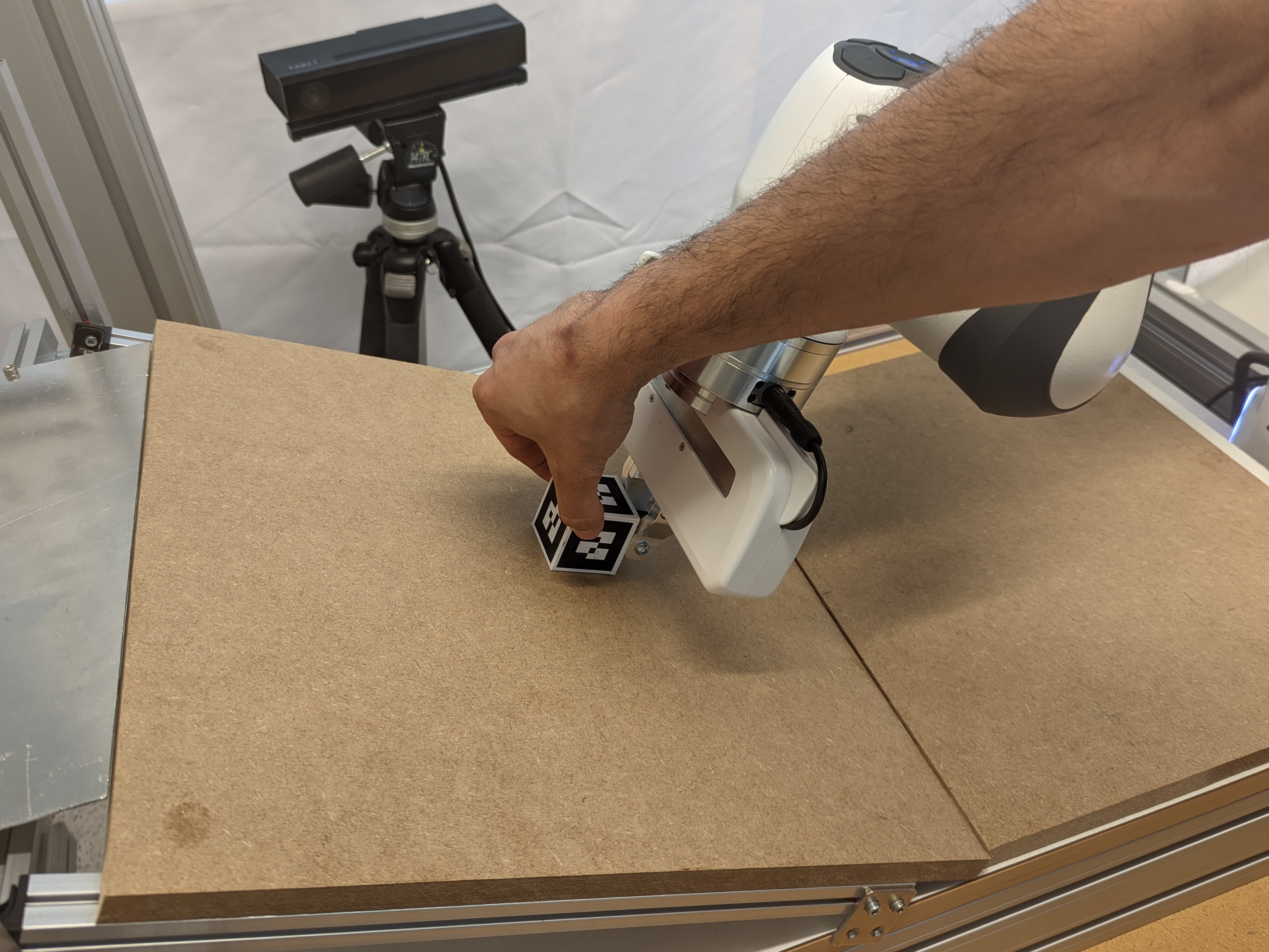} \hspace{0.025cm}
\includegraphics[height = 0.23\linewidth]{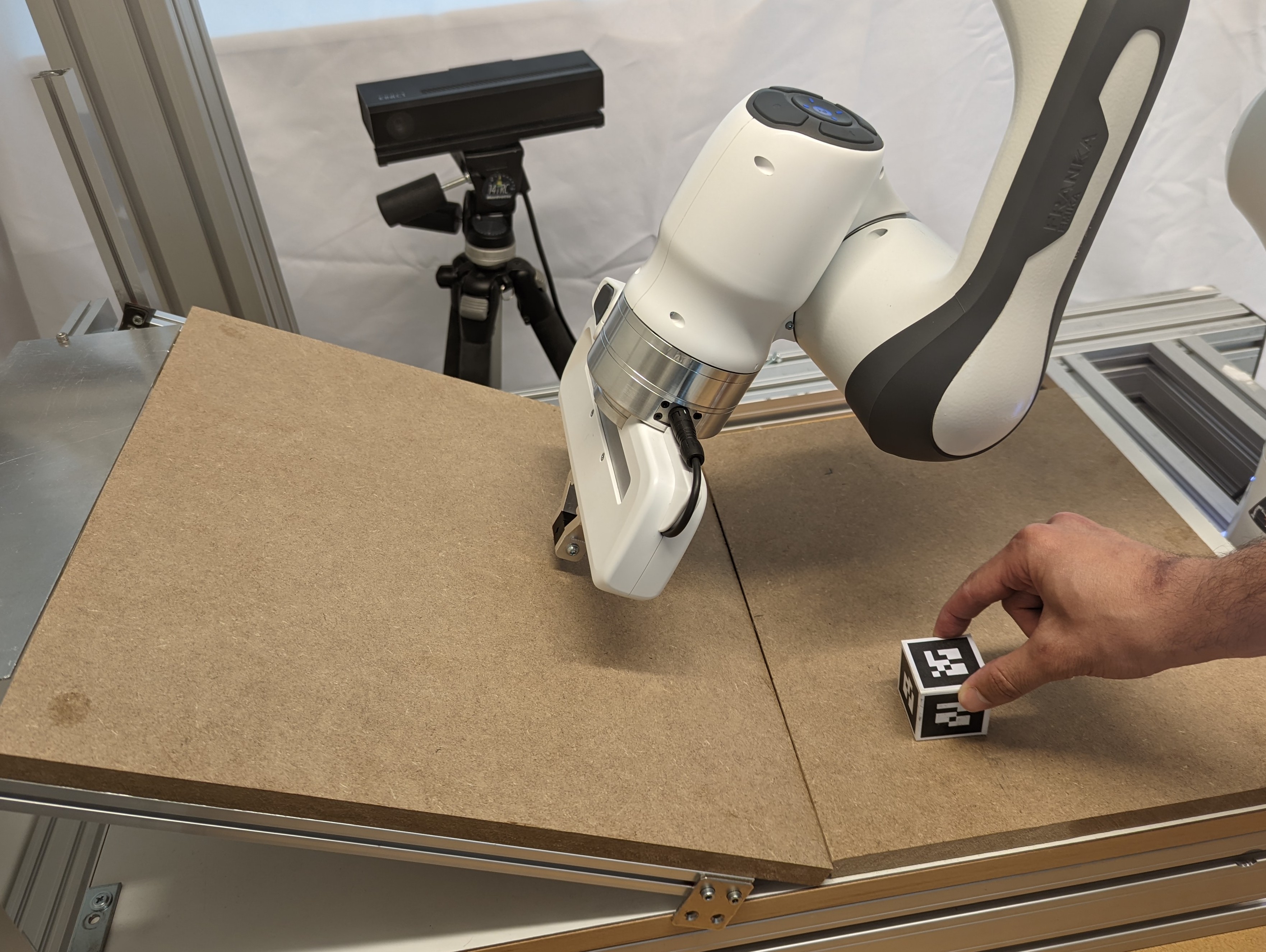} \hspace{0.025cm}
\label{fig:exp-react-global} 
}

%
%
\caption{Disturbances applied for reactivity experiments. To test SoT reactivity, in Fig. \ref{fig:exp-react-local} we randomly perturb the picking location of the cube while the robot is moving. To test BT reactivity, in Fig. \ref{fig:exp-react-global} we manually replace the cube from the ramp to a random position on the table.}
\label{fig:reactivity}
\end{figure}

In the first experiment, we evaluate the reactivity induced by the SoT framework by introducing a \textit{local} disturbance, in the sense that it does not result in a failure in the BT, but is rather handled online by the controller.
In particular, we randomly perturb the picking location of the cube while the robot is moving (Fig. \ref{fig:exp-react-local}).
This change is picked up by the Kinect camera and triggers an update on the goal location, which the SoT then conforms to.

In the second experiment, we evaluate the reactivity induced by the BT by introducing a \textit{global} disturbance, that is handled by the BT.
In particular, we simulate a failure during the pushing by manually replacing the cube from the ramp to a random position on the table (Fig. \ref{fig:exp-react-global}).
This artificially makes the condition node that checks that the cube is placed in front of the end-effector fail.
As a consequence the BT is forced to re-structure the SoT, and perform the full operation again.

Table \ref{tab:tab2} registers the success results for the combined 50 trials for both experiments.
The system successfully completed the task in 92\% and 88\% of the trials in case of \textit{local} and \textit{global} disturbances respectively.
For both experiments, all completed trials were performed at first attempt after the introduced disturbance.
As for the previous experiments, the failure rates are associated to inaccuracies in the tracking of the cube.
Four of the failed attempts (2 for \textit{local}, 2 for \textit{global}) happened during the placing of the cube, causing a collision similar to the one described in the robustness experiments in Section \ref{sec:robustness}.
The other registered failure happened during the picking, for which a tracking error caused the robot to close the gripper outside of the cube boundaries.

\begin{table}[h!]
\centering
\caption{Reactivity experiments}
\label{tab:tab2}
\begin{tabular}{lrr}
\toprule
Disturbance & \# Trials & Success rate \\
\midrule
\textit{Local}                & 25                 & 92\%                  \\
\textit{Global}                  & 25                 & 88\%				\\
\bottomrule
\end{tabular}
\end{table}

\section{Conclusion}
\label{sec:conclusion}
In this work, we combine a hierarchical Stack-of-Tasks control strategy with Behavior Trees for task composition in robot control.
The proposed framework benefits from the reactivity of the two models, without compromising on modularity and understandability.
The experiments on a 7-DOF manipulator show that the robot can robustly achieve a challenging goal, being able to react to unexpected changes.

A limitation of the current framework lies in the design of the hierarchical control problems in dynamic environments, as it becomes difficult to predict all possible scenarios in advance. Since recent works have showed promising results in the learning of BTs \cite{learning-colledanchise-2019, 8794104}, combining SoT with them opens new possibilities for learning how to successfully compose tasks.


\balance
\bibliographystyle{IEEEtran}
\bibliography{references}
\end{document}